\documentclass[journal]{IEEEtran}
\usepackage{amsmath,amsfonts}
\usepackage{algorithmic}
\usepackage{algorithm}
\usepackage{array}
\usepackage[caption=false,font=normalsize,labelfont=sf,textfont=sf]{subfig}
\usepackage{textcomp}
\usepackage{stfloats}
\usepackage{url}
\usepackage{verbatim}
\usepackage{graphicx}
\usepackage{cite}
\newcommand{\rv}[1]{\textcolor{black}{{#1}}}
\newcommand{\hk}[1]{\textcolor{black}{{#1}}}

\newcommand{\degree}[1]{°\ }
\DeclareUnicodeCharacter{2218}{\degree}
\usepackage{booktabs}
\usepackage{multirow}
\usepackage{xcolor,colortbl}
\newcommand{\gray}{\cellcolor{gray!20}}
\usepackage{mathtools}
\usepackage{graphicx} 
\usepackage{amssymb}

\usepackage{siunitx}
\usepackage{adjustbox}
\usepackage{subfig}

\begin{document}

\title{Cross360: 360° Monocular Depth Estimation via Cross Projections Across Scales}

\author{Kun~Huang,
        Fang-Lue~Zhang,~
        and~Neil~Dodgson
\IEEEcompsocitemizethanks{\IEEEcompsocthanksitem K. Huang, F.-L. Zhang, and N. Dodgson are from the School of Engineering and Computer Science, Victoria University of Wellington, New Zealand. This work was supported in part by the Marsden Fund Council managed by the
Royal Society of New Zealand under Grant MFP-20-VUW-180, and internal research grant (Project No. 400876) from Victoria University of Wellington.\protect\\
E-mail: \{kun.huang, fanglue.zhang, neil.dodgson\}@vuw.ac.nz
\IEEEcompsocthanksitem *Fang-Lue Zhang is the corresponding author.\protect\\}
\thanks{Manuscript received 2025; revised 2025.}}

\markboth{Journal of \LaTeX\ Class Files,~Vol.~14, No.~8, August~2021}%
{Shell \MakeLowercase{\textit{et al.}}: A Sample Article Using IEEEtran.cls for IEEE Journals}

\IEEEpubid{0000--0000/00\$00.00~\copyright~2021 IEEE}

\maketitle

\begin{abstract}
360° depth estimation is a challenging research problem due to the difficulty of finding a representation that both preserves global continuity and avoids distortion in spherical images. Existing methods attempt to leverage complementary information from multiple projections, but struggle with balancing global and local consistency. \hk{Their} local patch features have limited global perception, and the combined global representation does not address discrepancies in feature extraction at the boundaries between patches. To address these issues, we propose Cross360, a novel cross-attention-based architecture integrating local and global information using less-distorted tangent patches along with equirectangular features. Our Cross Projection Feature Alignment module employs cross-attention to align local tangent projection features with the equirectangular projection’s 360° field of view, ensuring each tangent projection patch is aware of the global context. Additionally, our Progressive Feature Aggregation with Attention module refines multi-scaled features progressively, enhancing depth estimation accuracy. Cross360 significantly outperforms existing methods across most benchmark datasets, especially those in which the entire 360° image is available, demonstrating its effectiveness in accurate and globally consistent depth estimation. The code and model are available at \url{https://github.com/huangkun101230/Cross360}.
\end{abstract}    


\section{Introduction}\label{sec:introduction}
A 360° image offers comprehensive environmental information by covering an entire field of view (FoV) of 180° $\times$ 360°. This extensive coverage has generated significant research interest in various areas, including depth estimation \cite{ai2024elite360d, li2022omnifusion, jiang2021unifuse, shen2022panoformer, ai2023hrdfuse, wang2020bifuse, wang2022bifuse++}, optical flow prediction \cite{li2022deep,yuan2021360,shi2023panoflow}, and surface normal estimation \cite{huang2024panonormal,karakottas2019360,huang2024multi}. The most common representation of a 360° image is the equirectangular projection \hk{(ERP)} \cite{miller1949equi}, which introduces significant distortions and presents challenges for understanding scene information \cite{su2017learning, eder2020tangent}.

Recent methods \cite{ai2024elite360d, li2022omnifusion, wang2022bifuse++} that adopt multiple projections for 360° depth estimation have shown promising results by exploring the complementary information provided by each projection. BiFuse \cite{wang2020bifuse} and HRDFuse \cite{ai2023hrdfuse} leverage local patches generated by cube map projection (CP) and tangent projection (TP) to learn regional structural features and use the ERP image to mitigate the spatial discontinuity of the patch-based features. However, one significant issue of these methods is only exploring information within the limited FoV of a patch when learning local features, which restricts the capability of understanding intricate scene details and leads to fragmented understanding. Although the ERP-based features provide a global view to combine local features, the descriptions of neighboring patches still lack continuity and coherence in the final representation. Existing padding \cite{wang2020bifuse, cheng2018cube} or fusion \cite{ai2023hrdfuse, li2022omnifusion} techniques were proposed to address this issue, but they often fail to achieve optimal performance.

In this paper, we introduce Cross360, a novel method that effectively integrates local and global information when learning scene features. We explore relationships between the patches of the less-distorted tangent projection and the global scene represented by the equirectangular projection. The ERP features, with their full (180° $\times$ 360°) FoV, provide a holistic understanding at a coarse level and identify broad patterns and relationships. Conversely, the TP patches in the limited-FoV focus on local finer regions, capturing detailed patterns and textures. To ensure each TP patch is aware of global contexts and scales, we propose a Cross Projection Feature Alignment (CPFA) module that employs a cross-attention mechanism to align the limited-FoV local TP patch features with the full-FoV ERP representations. 
\hk{Unlike prior ERP/TP fusion frameworks that simply concatenate or self-attend within a single projection, our architecture performs multi-level cross-projection alignment at every decoder stage. CPFA explicitly exchanges information between TP and ERP features so that each TP patch incorporates the entire ERP context, enabling the learning of both local and global receptive representations and ensuring smooth transitions and continuity across projections.}
Furthermore, we introduce a Progressive Feature Aggregation with Attention (PFAA) module, which aggregates multi-scale features from small to large using attention maps. This module progressively refines the final representation and predicts the 360° depth map,
\hk{allowing Cross360 to attend to various levels of detail and capture complex visual information across multiple scales. In addition, the slightly overlapping TP patch layout together with bidirectional ERP–TP transformations keeps neighboring patches geometrically consistent, which effectively alleviates boundary artifacts. Together, these components form a hierarchical architecture that delivers superior global consistency and significantly improves depth estimation accuracy.}

\IEEEpubidadjcol

We conduct extensive experiments on four commonly used datasets: Matterport3D \cite{chang2017matterport3d}, Stanford2D3D \cite{armeni2017joint}, Structured3D \cite{zheng2020structured3d} and 3D60 \cite{zioulis2018omnidepth}, evaluating our method against the latest depth estimation approach for both perspective and 360° images. Our approach substantially exceeds the performance of current state-of-the-art (SOTA) methods across most datasets, especially when the full regions of 360° images are available. Our main contributions are:
\begin{enumerate}
    \item We present Cross360, a 360° monocular depth estimation method that addresses distortion and discontinuity issues through cross-attention across projections and scales. Our approach achieves state-of-the-art performance with the complete 360° images are available.
    \item We propose a CPFA module that utilizes a cross-attention mechanism to align limited-FoV local patch (TP) features with full receptive fields (ERP), enabling the learning of local and global receptive representations across both ERP and TP domains.
    \item We introduce a PFAA module that aggregates multi-scale features from the CPFA module using attention maps, refining the final representation and outputting the 360° depth predictions.
\end{enumerate}
\section{Related Work}\label{sec:relatedwork}
\hk{We review prior work in two categories.
The first category, 360° depth estimation, encompasses methods designed to predict depth from various 360° projection formats.
These studies differ in how they handle the intrinsic challenges of 360° imagery, such as spherical distortion, seam discontinuities between patches, and global context modeling.
The second category, cross-attention mechanisms, covers approaches that align or fuse features across distinct domains or scales.
These techniques provide the foundation for our proposed CPFA module, which requires effective interaction between tangent patch and ERP features.}
\subsection{360° Depth Estimation}
The monocular 360° depth estimation task typically uses 360° images in the ERP format as input, capturing omnidirectional information for the entire scene. However, \hk{ERP images} present two well-known challenges: distortion and discontinuity. To address these issues, researchers have explored various approaches.

Since conventional \hk{convolutional neural network (CNN)} kernels are designed for the perspective images and are inefficient at overcoming distortion, several methods \cite{artizzu2023omni,li2023spherical,wu2020spherical,tateno2018distortion} have focused on mitigating severe distortion by proposing distortion-aware kernels, and algorithms  \cite{chen2021distortion,cheng2020omnidirectional} further apply the deformable convolutional network to learn an offset of the spherical distortion. These kernels adjust the receptive fields of the CNN according to the latitude of the sphere when extracting features. A previous work \cite{li2021omnidirectional} adopts such spherical kernels for stereo omnidirectional depth estimation tasks via the linear epipolar constraint on the spherical surface. While some methods transform spherical maps to other domains using schemes such as pixelization \cite{shakerinava2021equivariant, perraudin2019deepsphere} or the icosahedral polyhedron representation \cite{Yoon_2022_CVPR, Lee_2019_CVPR}, incorporating corresponding CNN feature extraction strategies, a recent study \cite{liu2024estimating} focuses on the sphere surface. It proposes an independent spherical kernel for each point based on its eight equidistant neighbors on the closest outer circle on the sphere to preserves geometric properties and minimizes the discontinuity issue.

Many algorithms estimate 360° depth maps through various projection techniques. For example, PanoDepth \cite{li2021panodepth}, HoHoNet \cite{sun2021hohonet}, and ACDNet \cite{zhuang2022acdnet} use the ERP image as input and aggregate features to predict the 3D scene. SliceNet \cite{pintore2021slicenet} divides the ERP image vertically on the sphere to investigate the spatial relationships among slices without employing a transformer mechanism. In contrast, PanelNet \cite{yu2023panelnet} incorporates geometric embedding with multi-head self-attention, and EGFormer \cite{yun2023egformer} applies attention to both horizontal and vertical slices to enhance the network's ability to capture spatial dependencies across different directions. To address the distortion issue, SalNet360 \cite{monroy2018salnet360} converts the ERP input with CP for dense predictions on less distorted surfaces, then unfolds them back to the ERP domain. Following the work \cite{eder2020tangent} of transferring the ERP image with TP, 360MonoDepth \cite{rey2022360monodepth} and OmniFusion \cite{li2022omnifusion} investigate depth prediction on tangent patches and fuse them back to the complete ERP image. PanoFormer \cite{shen2022panoformer} uses a U-shaped transformer \cite{wang2022uformer} to find global dependencies among the tangent patches. While some works \cite{cheng2018cube} propose CP image padding methods to improve connectivity between faces, other algorithms such as GLPanDepth \cite{bai2024glpanodepth}, BiFuse++ \cite{wang2022bifuse++}, UniFuse \cite{jiang2021unifuse}, and BiFuse \cite{wang2020bifuse} incorporate a second image projection format (ERP) with CP, and HRDFuse \cite{ai2023hrdfuse} combines TP for gathering more comprehensive global information, addressing discontinuity among patches and fine-grained details within the non-distorted patches simultaneously. Recently, Elite360D \cite{ai2024elite360d} fuses the icosahedron geometric information to provide spherical information, and \hk{SGFormer \cite{zhang2025sgformer} integrates spherical geometric priors into vision transformers through bipolar reprojection and curve-local embedding.}
\hk{Other works \cite{liu2024estimating, wang2024depth, cao2025panda} adopt teacher–student frameworks to exploit large sets of pseudo-labeled data. PanDA \cite{cao2025panda} adapts Depth Anything V2 \cite{depthanything, yang2024depth} to spherical imagery and achieves state-of-the-art results, but it requires about 120k 360° images (20k labeled and 100k unlabeled), making a direct comparison with our method unfair due to the much larger training set. DepthAnyWhere \cite{wang2024depth} follows a similar teacher–student strategy using Depth Anything \cite{depthanything} to generate pseudo labels and likewise depends on substantially more pseudo-labeled data than the ground-truth annotations in our benchmarks, so a direct metric comparison would also be inequitable.}
\hk{Unlike these teacher–student approaches that rely on massive pseudo-labeled datasets,} Cross360 utilizes both ERP and TP projections, each in its own format, and employs cross-attention to align features between these projections. \hk{This strategy captures detailed local information from the TP patches while simultaneously modeling global dependencies from the ERP representation, allowing us to learn robust and geometry-aware features without requiring additional large-scale pseudo supervision.}

\subsection{Cross-attention}
Cross-attention is widely used to capture efficient features across different patch scales or feature domains. For instance, CrossVit \cite{chen2021crossvit} introduces two branches that simultaneously produce small-patch and large-patch tokens, learning a multi-scale representation. Similarly, CAT \cite{lin2022cat} employs hierarchical attention between patches and within each patch. Furthermore, CCNet \cite{huang2019ccnet} proposes a lightweight transformer that aggregates global dependencies pixel-wise by utilizing information from each pixel in both horizontal and vertical directions. Additionally, AutoAlign \cite{chen2022autoalign} introduces a cross-attention feature alignment module to learn a map that aligns features between 2D images and 3D point clouds for 3D object recognition. Recently, $S^2$Net \cite{li2023mathcal} applies the Swin transformer \cite{liu2021swin} and HEALPix \cite{gorski2005healpix} to sampling pixels for extracting features from the ERP domain and models global dependencies with features from the spherical domain. In our approach, we introduce the CPFA module to establish relationships between each tangent patch that is derived from the ERP image, and the extracted ERP features. The CPFA module enables each tangent patch to gain a comprehensive understanding of the global context, while still preserving the local details within the image. This dual capability ensures that each patch can seamlessly integrate holistic image information without losing the specific characteristics of either local or global regions.

\begin{figure*}[t]
 \includegraphics[width=1\linewidth]{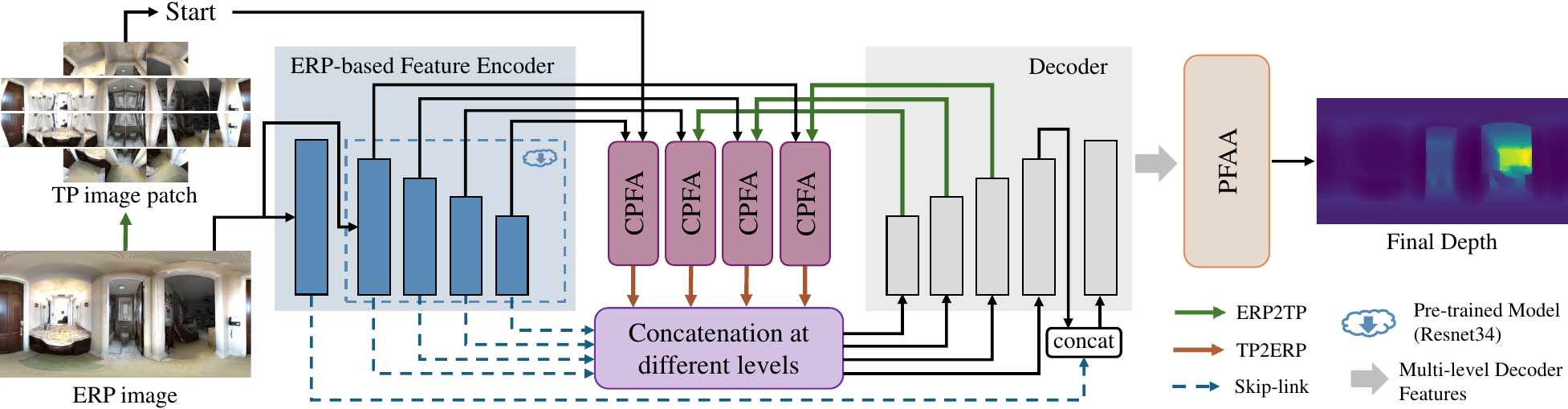}
  \caption{Framework Overview: Cross360 estimates 360° depth through three main components: an ERP-based feature encoder, Cross Projection Feature Alignment (CPFA), and Progressive Feature Aggregation with Attention (PFAA). The CPFA module aligns TP and ERP features, which are then concatenated with skip-linked ERP features and passed to the decoder. The decoder generates features at each level, used for multi-scale supervised learning and passed to the next CPFA level. The PFAA module processes all decoder features to produce the final depth map. Note that, multi-scale depth predictions during the decoding stage are not shown for better visualization.
}
 \label{fig:network}
\end{figure*}
\section{Methodology}\label{sec:method}

\subsection{Overview}

The framework of Cross360 is illustrated in \hk{Fig. \ref{fig:network}}, which takes an ERP image and a set of TP images (converted from the ERP image) as inputs, and outputs the predicted depth map. It comprises three main components: ERP-based feature encoder, CPFA, and PFAA. Our goal is to effectively utilize the complementary information from both local and global feature representations at varying scales, which is crucial for accurate 360° depth estimation with minimized distortion and discontinuity issues.

The ERP-based feature encoder takes an ERP image 
and extract hierarchical feature representations $F^{ERP}_{s}$ across different scales from $s=1$ to $S$, where $s = S$ is the finest resolution (by default, we set $S=5$).
To mitigate distortion in the ERP image, CPFA modules are leveraged to align feature representations from the TP patches and ERP images. As shown in \hk{Fig. \ref{fig:network}}, the process involving TP images begins by feeding the patches into the coarsest-scale CPFA module. At this scale, the CPFA module applies self-attention to a set number, $N$ (by default, we set N=26), of the embedded TP patches, ensuring that the local feature maps $\{F^{TP}_{n}\}_{n=1,..., N}$ are well-represented and contextualized without distortion. Next, CPFA employs a cross-attention mechanism to align the TP features with the global context from the ERP features at the corresponding scale, enhancing local feature maps to capture complementary structural information. The aligned cross-projection feature map $F^{CA}_{s}$ is then transformed into the ERP domain, concatenated with the skip-linked ERP features, and fed into the convolutional decoder at scale $s$. The decoded feature $\{F^{D}_{s}\}_{s = 1,..., S-2}$ is subsequently processed by the CPFA module of the next scale. Notably, except for the coarsest and finest scale, the CPFA module at scale $s$ takes the tangent patches of $F^{CA}_{s-1}$ to calculate cross-attention with $F^{ERP}_{s}$. The decoded features across all scales, $\{F^{D}_{s}\}_{s = 1,..., S-1}$, are also used to predict depth maps at various scales for multi-scale supervision. At the finest scale of the decoder, it takes the previous scale's feature and $F^{ERP}_{s=S}$ to generate $F^{D}_{s=S}$. The decoded features $\{F^{D}_{s}\}_{s = 1,..., S}$ then proceed to the PFAA module, which progressively integrates the most relevant information from different scales, ultimately producing the final depth prediction results at the finest resolution. The details are as follows.

\subsection{ERP-based Feature Encoder}
The ERP-based feature encoder extracts features $F^{ERP}_{s}\in \mathbb{R}^{c\times h\times w}$ at each scale $s$ from the input ERP image of size $3 \times H \times W$, where $h=H/2^{S-s},w=W/2^{S-s}$, and \hk{$c$} is the number of feature channels. Our encoder comprises a convolutional block for extracting fine-grained details at the finest resolution of $H \times W$ and a pre-trained ResNet34 \cite{He_2016_CVPR} block for features at smaller scales. Compared to using ResNet34’s pooling layers to downsample the original scale, the added convolutional block avoids losing the finest-level appearance information. With the ERP encoder, the 360° input is treated as a single, unified entity to capture features, compensating for better continuous details rather than distortion-free but patchified information. This also allows our encoder to incorporate any existing pre-trained backbones, offering flexibility and potential performance improvements with popular models. However, these pre-trained 2D perspective backbones still suffer from spherical distortions (see DepthAnything and ASNGeo in Table \ref{tab:quantitative_comparison}), and thus, can only be used to provide the initial feature representations. 

\subsection{Cross Projection Feature Alignment}
\begin{figure}[t]
 \includegraphics[width =1\linewidth]{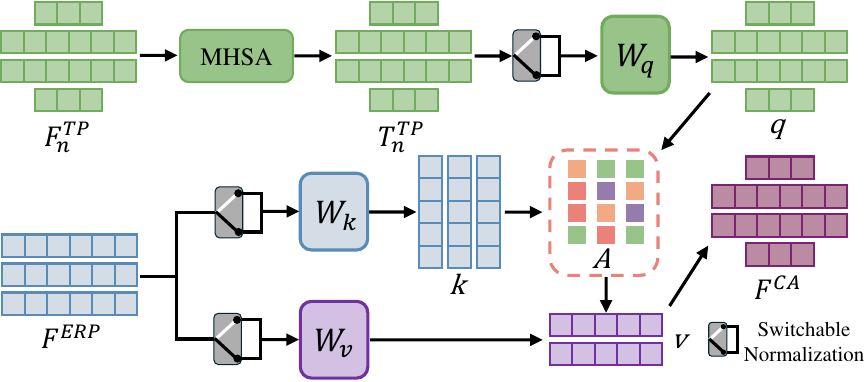}
 \caption{The architecture of Cross Projection Feature Alignment module.}
 \label{fig:cpfa_module}
\end{figure}

We employ several CPFA modules in our network at various scales to align TP-based features with ERP-based features via cross-attention. At each level, the aligned feature map $F^{CA}_{s}$ is then converted to ERP format and concatenated with the corresponding ERP features $F^{ERP}_{s}$ for subsequent decoding. The decoded feature $F^{D}_{s}$ is then upsampled and transformed back into TP patches for cross-projection alignment at the next scale. A CPFA module's structure is shown in \hk{Fig. \ref{fig:cpfa_module}.}
\subsubsection{\hk{ERP2TP and TP2ERP}}
\hk{Projecting 360° inputs onto tangent planes minimizes spherical distortion while preserving geometric properties. The ERP2TP transformation converts equirectangular images to perspective-like patches by projecting spherical coordinates onto tangent planes \cite{li2022omnifusion,ai2023hrdfuse,eder2020tangent}. The inverse TP2ERP transformation maps processed patches back to the spherical format using inverse spherical projection and interpolation.}
Following OmniFusion \cite{li2022omnifusion}, we use tangent patches transformed from the ERP format via gnomonic projection \cite{coxeter1961introduction}. We also use a non-uniform patch sampling strategy on the ERP image to ensure equal coverage of the spherical content. Therefore, we sample fewer patches near the poles to prevent excessive overlap. We use a set of $N=26$ tangent patches for a balance of speed and performance (refer to our ablation study). Tangent patches are sampled at five different latitudes: $\num{-72}$°, $\num{-36}$°, $\num{0}$°, $\num{36}$°, $\num{72}$°, with 3, 6, 8, 6, 3 patches respectively.
\hk{All tangent patches have a uniform 72° FoV, but each patch is sampled at four different resolutions—4, 8, 16, and 24 pixels in height and width—creating a multi-resolution hierarchy. The $4 \times 4$ resolution corresponds to the image domain, while the 8, 16, and 24 levels operate in the feature space.}

\subsubsection{Self-attention among Tangent Patches}
After sampling a set of TP patches, we process them through a multi-head self-attention (MHSA) block \cite{vaswani2017attention}. The MHSA enables the model to effectively capture contextual relationships among the tangent patches, leveraging the strengths of self-attention to focus on local details within the patches while facilitating interactions between different patches. The output feature $T^{TP}_{n}$ from this block will be used as the query for a subsequent cross-attention module.

\subsubsection{Cross-attention among Projections}
Previous methods such as BiFuse \cite{wang2020bifuse} and UniFuse \cite{jiang2021unifuse} integrate feature maps between ERP and CP by transforming the ERP image into CP patches or vice versa, based on their geometric relationships. OmniFusion \cite{li2022omnifusion}, in contrast, uses a self-attention mechanism to capture intricate information among TP patches. While these methods effectively achieve pixel-wise feature matching across projection domains, their performance is limited by two main issues: First, each geometry-transformed CP or TP patch has a limited FoV, lacking global perception. Second, these projections focus on detailed information within each patch without accounting for neighboring patches, leading to a lack of continuity. Although some padding or fusion techniques have been proposed to address this issue, the performance remains constrained.

To address these issues, our CPFA module leverages precise geometric transformations to match ERP and TP feature maps while overcoming the aforementioned limitations. As illustrated in \hk{Fig. \ref{fig:cpfa_module}}, after obtaining the ERP features \( F^{ERP}_{s} \) and the TP features \( T^{TP}_{n} \) from the self-attention block, we use the Switchable Normalization module \cite{SwitchableNorm} to learn the weights of channel-wise, layer-wise, and batch-wise normalization, flexibly selecting the most effective strategy for different components and the information from different projections. Note that, only at the coarsest scale are TP images directly fed into the CPFA. For each of the subsequent scales, the aligned feature map from the previous scale, initially in ERP format, is transformed into tangent patches for cross-attention (CA) at the current scale. Given the normalized feature maps, we generate the query \( q \) from each patch-wise feature \( T^{TP}_{n} \) and produce the key \( k \) and value \( v \) from the entire ERP feature map \( F^{ERP}_{s} \) with corresponding resolution sizes. We then calculate the attention map \( A \) based on \( q \) and \( k \), and obtain the projection-aligned feature with \( F^{CA} = A \cdot v \). This process can be formulated as:
\begin{equation}
\begin{array}{c}
q = T^{TP}_{n}\cdot\textbf{W}_{q}, \quad k = F^{ERP}\cdot\textbf{W}_{k}, \quad v = F^{ERP}\cdot\textbf{W}_{v} \\[0.10cm]
\scalebox{0.96}{$A = \text{softmax}(q\cdot k^{T}), \quad F^{CA} = CA(F^{TP}_{n}, F^{ERP}) = A\cdot v$}
\end{array}
\end{equation} 
where $\textbf{W}_{q}, \textbf{W}_{k}, \textbf{W}_{v}$ are learnable parameters. Note that generating the attention map in our cross-attention is more efficient than in all-attention, as the computation and memory complexity are linear rather than quadratic.
\subsubsection{Decoder}

At each scale (except the finest scale), the output of the CPFA module is concatenated with the skip-linked ERP features and processed by the corresponding layer of our decoder to produce the decoded feature maps, denoted as $\{F^{D}_{s}\}_{s = 1,\dots,S-1}$. This concatenation combines both the skip-linked ERP input and the CPFA output to facilitate comprehensive feature representation for the entire image. The resulting feature maps are upsampled and projected from the ERP domain to tangent planes, generating tangent feature map patches that are then fed into the CPFA module at the next scale.

For the finest scale, we directly use the concatenation of the previous decoder's feature map $F^{D}_{s-1}$ and the finest ERP feature map $F^{ERP}_{s=S}$ as input, producing the feature map for this final resolution $F^{D}_{s=S}$. The stacked decoded features $\{F^{D}_{s}\}_{s = 1,...,S}$ across multiple scales effectively integrate global and local geometry and appearance features. These feature maps are subsequently passed to the final feature aggregation stage to predict the full-resolution depth map and are also used for multi-scale supervision by presenting depth maps at different scales through output layers.

\begin{figure}[t]
 \includegraphics[width =1\linewidth]{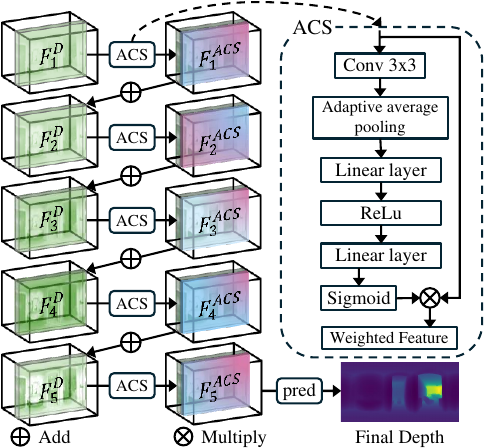}
 \caption{The architecture of Progressive Feature Aggregation with Attention module.}
 \label{fig:pfaa_module}
\end{figure}

\subsection{Progressive Feature Aggregation with Attention}
The PFAA module takes the set of multi-scaled decoded features $\{F^{D}_{s}\}_{s = 1,...,S}$ from the decoders, gradually aggregating features across scales to refine and enhance the final representation based on the most relevant features at each scale, and outputs the final depth map estimation. Attention maps help the model progressively integrate selected information from different scales, thereby improving performance. As shown in Fig. \ref{fig:pfaa_module}, the attention maps $\{F^{ACS}_{s}\}_{s = 1,...,5}$ for the decoded features of each scale are calculated via an Attention Cross Scale (ACS) block, $\mathcal{F}_{s}$. The process begins with the smallest scale feature map, where the input feature map for the next scale is derived solely from the ACS block. Subsequently, the attention information is added to the features of the next scale, serving as the input for the following ACS block. This approach adaptively refines features by emphasizing relevant channels, resulting in more effective integration of information across different scales and ultimately enhancing performance. This approach is detailed in the following formulation:
\begin{equation}
\begin{array}{c}

F^{ACS}_{1}=\mathcal{F}_{1}(F^{D}_{1}) \cdot F^{D}_{1} \\[0.1cm]
F^{ACS}_{s}=\mathcal{F}_{s}(F^{D}_{s}+F^{ACS}_{s-1}) \cdot (F^{D}_{s}+F^{ACS}_{s-1}), s=2,\dots,S
\end{array}
\end{equation} 

\subsection{Optimization}
Due to Cross360's capability to effectively learn geometric and appearance features across the spherical domain, we use the output features from the decoder $\{F^{D}_{s}\}_{s = 1,...,S-1}$ and PFAA modules $F^{ACS}_{s=S}$, passing them to the specific convolutional layers (Conv) to predict depth maps $D_{s}$ at various scales, which can be formulated as follows:

\begin{equation}
\begin{array}{c}
D_{s}=\text{Conv}(F^{D}_{s}), s=1,\dots,S-1\\[0.1cm]
D_{s}=\text{Conv}(F^{ACS}_{s}), s=S
\end{array}
\end{equation}

For the Matterport3D, Structured3D, and 3D60 datasets, our experimental results show that the following settings are more effective. We employ a combination of MSE loss $\mathcal{L}_{mse}$ and gradient loss $\mathcal{L}_{grad}$ for multi-scale pixel-wise depth supervision with ground truth $\hat{D}_{s}$, which are defined as:
\begin{equation}
\mathcal{L}_{mse} = \sum\limits_{s=1}^{S}\sum\limits_{i=1}^{N} \| D^{i}_{s} - \hat{D}^{i}_{s} \|_2
\end{equation}

\begin{equation}
\mathcal{L}_{\text{grad}} = \sum\limits_{s=1}^{S}\frac{1}{N} \sum\limits_{i=1}^N \left( \left| |\nabla D_{\text{s}}^i| - |\nabla \hat{D}_{\text{s}}^i| \right| \right),
\end{equation}
where $N$ is the total number of pixels, and $\nabla$ represents the gradient operations implemented as convolutions with the Sobel kernels.
The total loss $\mathcal{L}_{total}$ is:
\begin{equation}
\mathcal{L}_{total} = \mathcal{L}_{mse} + \mathcal{L}_{grad} .
\end{equation}
For Stanford2D3D dataset, which contains significantly less training data, we follow previous works \cite{ai2024elite360d,shen2022panoformer} and adopt the reverse Huber loss \cite{laina2016deeper} $\mathcal{L}_{BerHu}$ to supervise multi-scale pixel-wise depth loss and gradient loss.

\section{Experiments and Results}\label{sec:results}
We validate our method using four benchmark datasets: two real-world datasets, Matterport3D (M3D)\cite{chang2017matterport3d}, Stanford2D3D (S2D3D) \cite{armeni2017joint}, and two synthetic datasets, Structured3D (Struct3D)\cite{zheng2020structured3d} and 3D60 \cite{zioulis2018omnidepth}. We compare our method with SOTA approaches, including methods designed for perspective images, DepthAnything \cite{depthanything} and ASNGeo \cite{long2024adaptive}, and 360° depth estimation methods. \rv{For real-world benchmarks (Table \ref{tab:quantitative_comparison_clip}), we report results following established evaluation protocols (e.g., UniFuse, PanoFormer and Elite360D) to ensure direct comparability with published SOTA numbers. In contrast, for synthetic datasets (Table \ref{tab:quantitative_comparison}), we retrain all baseline methods from scratch using official code and hyperparameters to guarantee a fair comparison under identical conditions.} Additionally, we conduct an ablation study to evaluate key components of our approach.

\begin{figure*}
    \centering
    \subfloat[\footnotesize Original TP Sampling (N=26)]{
        \includegraphics[width=0.45\textwidth]{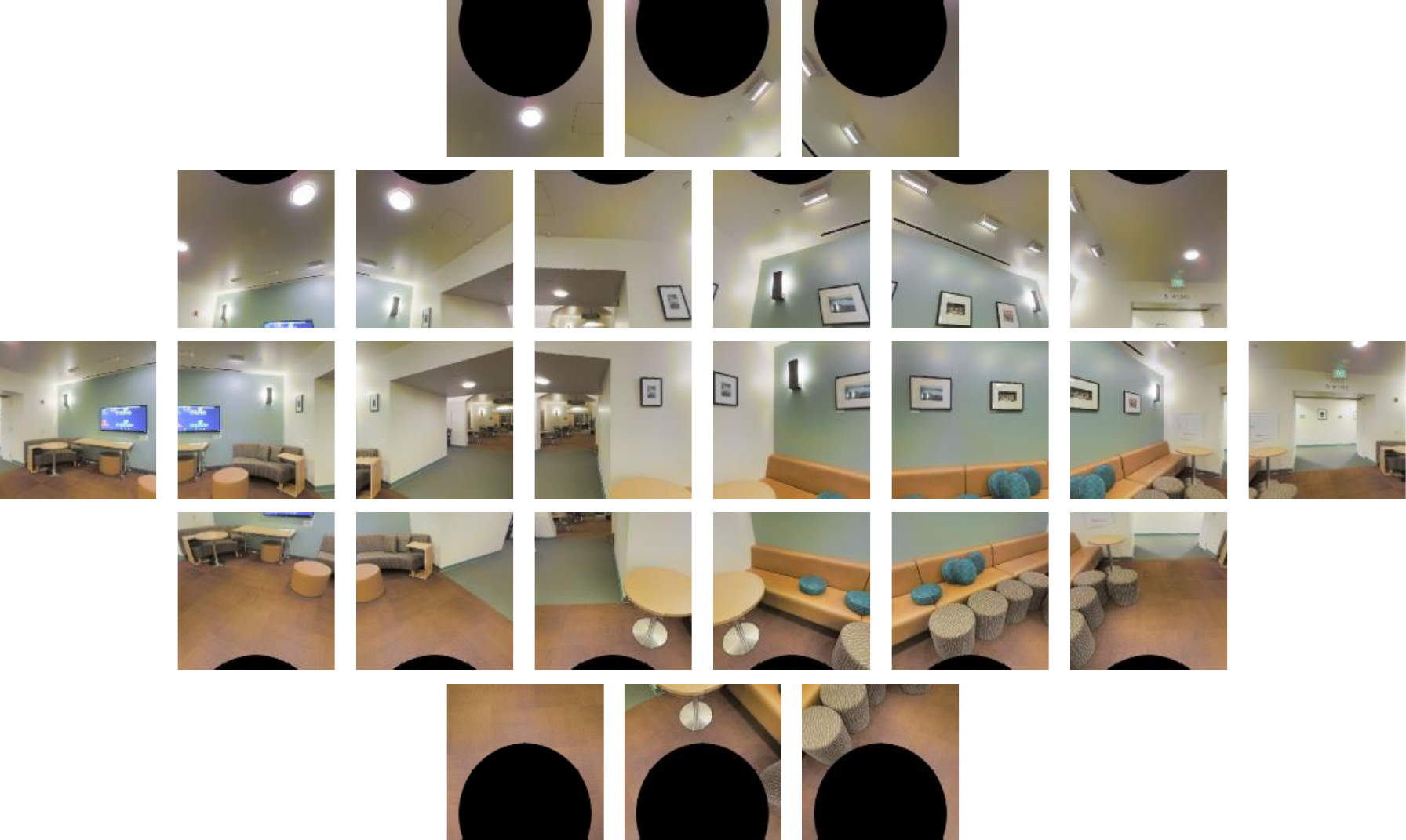}
        \label{fig:ori_sampling}
    }
    \hfill
    \subfloat[\footnotesize Original ERP input]{
        \includegraphics[width=0.45\textwidth]{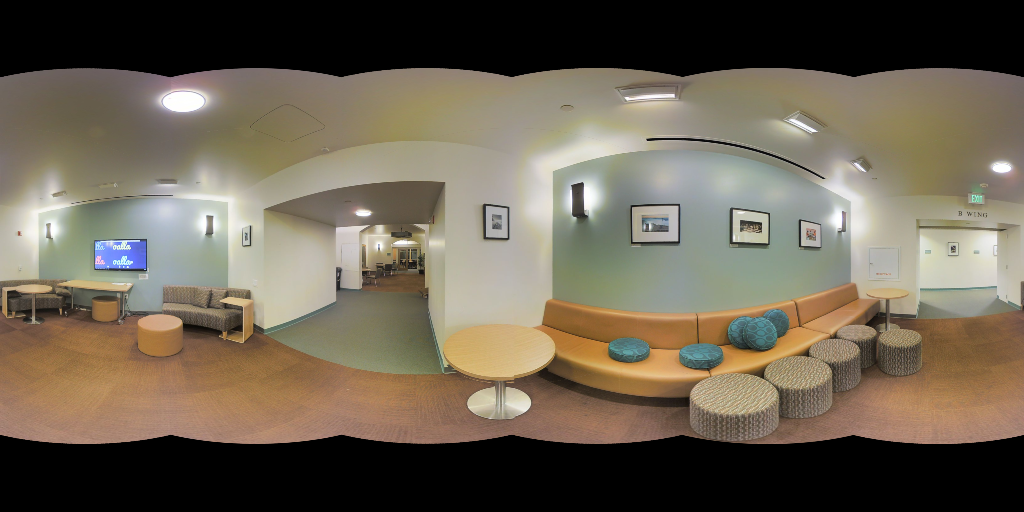}
        \label{fig:ori_input}
    }
    \caption{Real-world data input with the original TP sampling strategy}
    \label{fig:original_inputs}
\end{figure*}

\begin{figure*}
    \centering
    \subfloat[\footnotesize Specific TP Sampling (N=20)]{
        \includegraphics[width=0.45\textwidth]{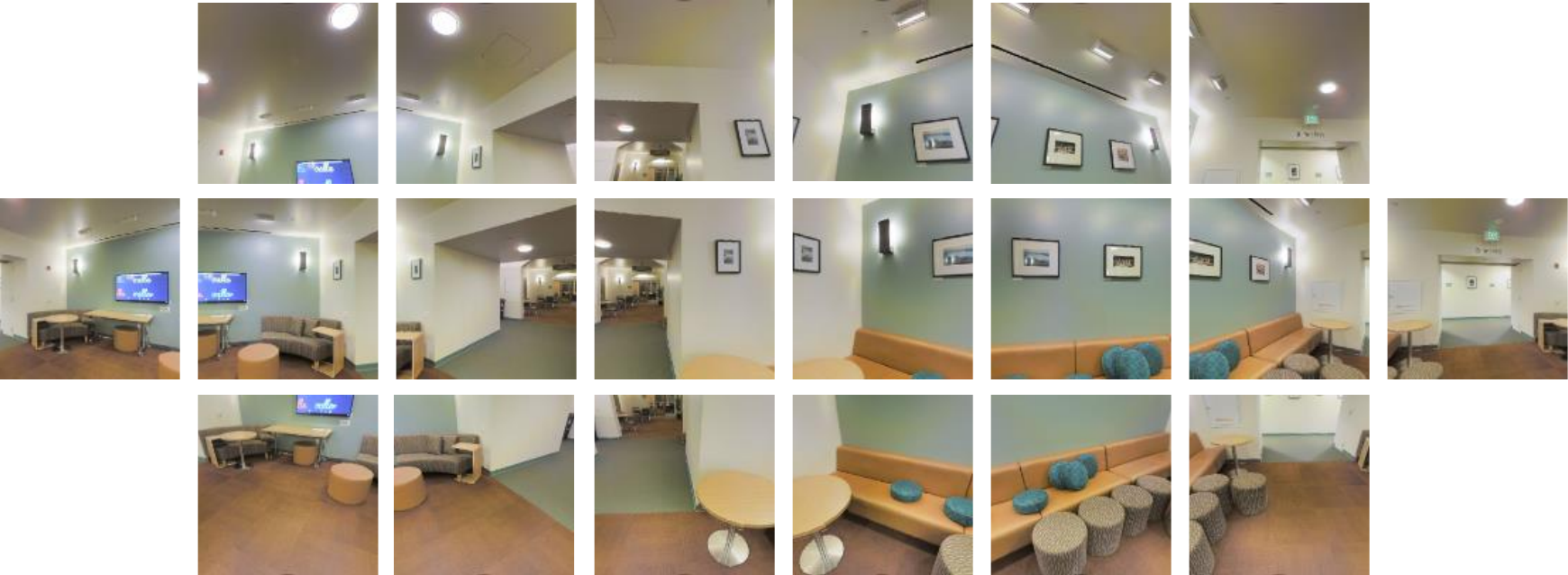}
        \label{fig:clip_sampling}
    }
    \hfill
    \subfloat[\footnotesize Clipped ERP input]{
        \includegraphics[width=0.45\textwidth]{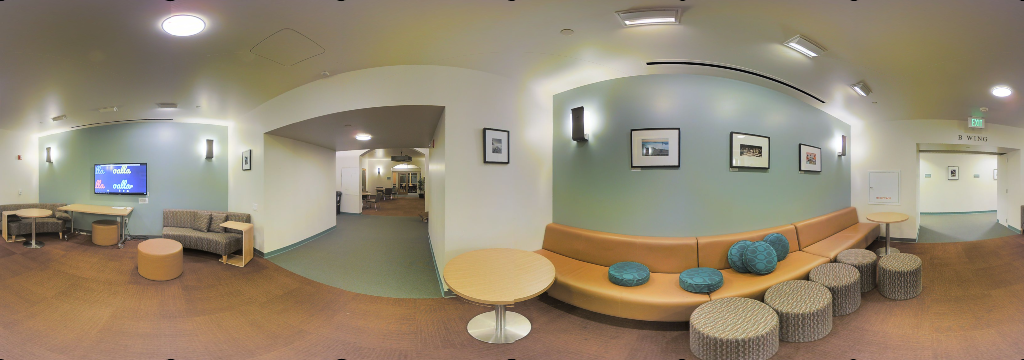}
        \label{fig:clipped_input}
    }
    \caption{Real-world data input with the specific TP sampling strategy}
    \label{fig:clipped_inputs}
\end{figure*}

\subsection{Evaluation Metric, Datasets and Hardwares}
We evaluate depth estimation performance using three error metrics: absolute relative error (Abs Rel), Squared Relative Error (Sq Rel) and Root Mean Square Error (RMSE) along with three accuracy metrics. These accuracy metrics measure the percentage of pixels where the ratio ($\delta$) between the predicted depth and ground truth is less than $1.25^1$, $1.25^2$, and $1.25^3$. For the real-world datasets, Matterport3D and Stanford2D3D, we follow the experimental settings used in previous works \cite{jiang2021unifuse, shen2022panoformer, ai2024elite360d}. \hk{All methods on these two datasets are evaluated using the datasets' standard test ERP images, which inherently contain missing regions at the top and bottom. We provide every method with the same masked ERP inputs without any padding or inpainting, and compute all metrics strictly on the valid pixels specified by the dataset’s ground-truth masks.} For the synthetic datasets, Structured3D and 3D60, we retrain previous models under their respective settings to further evaluate their performance, particularly in handling the most distorted regions in the ERP format.
\hk{All experiments are conducted under consistent experimental settings for fairness, and we performed our method using a single Intel Xeon W-2133 CPU core and an NVIDIA RTX 3090 GPU.}

\subsection{Training Details on Real-world Datasets}

Matterport3D and Stanford2D3D are widely used real-world 360° ERP datasets, offering high-resolution and detailed indoor data. However, both datasets contain missing regions at the top and bottom of images due to camera limitations restricting the vertical field of view or the deliberate exclusion of irrelevant noise. While some existing methods \cite{jiang2021unifuse, wang2020bifuse} are minimally affected by this vertical incompleteness, our CPFA module is more sensitive to it. The CPFA module identifies interactions between TP patches and aligns them with the global ERP context using cross-attention. When the ERP image has black regions at the poles, the network may learn ineffective features, disrupting the global geometric relationships between TP patches and ERP regions.

To address this issue, we modify the extracted TP and ERP feature maps by removing invalid regions, ensuring that the network focuses on meaningful information. Specifically, for datasets with complete fields of view, we sample \(N = 26\) TP patches across five latitudes (\(-72^\circ, -36^\circ, 0^\circ, 36^\circ, 72^\circ\)) with 3, 6, 8, 6, and 3 patches per row, respectively (as shown in Fig. \ref{fig:ori_sampling}). However, for datasets with missing regions at the poles, we adjust the sampling strategy by excluding affected latitudes and modifying patch distributions. In such cases, patches are sampled at \(-31.2^\circ, 0^\circ, 31.2^\circ\), with 6, 8, and 6 patches per row, respectively (Fig. \ref{fig:clip_sampling}). Additionally, we rotate the sampled TP patch centers away from the missing regions to maximize valid data utilization.

These modifications allow our CPFA module to maintain effective global feature alignment despite vertical incompleteness. Table \ref{tab:quantitative_comparison_clip} presents quantitative results demonstrating the impact of these adjustments on real-world datasets. Furthermore, to validate our approach under fully covered ERP inputs, we conduct additional comparisons using synthetic datasets, which provide complete environmental information, particularly in the most distorted ERP regions (Table \ref{tab:quantitative_comparison}).

\begin{table*}[t]
    \centering
    \caption{Quantitative comparisons against SOTA methods on real-world datasets. Inputs cover incomplete FoV along the vertical direction.}
    \resizebox{\textwidth}{!}{
    \begin{tabular}{c|l|c|c|ccc|ccc}
        \toprule
        \multirow{1}{*}{\centering Dataset} & \multirow{1}{*}{Method}& \multirow{1}{*}{Pub'Year}& \multirow{1}{*}{Params (M)}
        &Abs Rel $\downarrow$  & Sq Rel $\downarrow$ & RMSE $\downarrow$
        & $\delta_1$(\%) $\uparrow$& $\delta_2$(\%) $\uparrow$& $\delta_3$(\%) $\uparrow$\\
        \midrule

        \multirow{9}{*}{\centering M3D}
         & EGFormer \cite{yun2023egformer} &ICCV'23 &15.39 &0.1473	&0.1517	&0.06025	&81.58	&93.90	&97.35\\
         & PanoFormer \cite{shen2022panoformer} &ECCV'22 &20.38 &0.1051	&0.0966	&0.4929	&89.08	&96.23	&98.31\\
         & BiFuse \cite{wang2020bifuse} &CVPR'20 &56.01  &0.1126	&0.0992	&0.5027	&88.00	&96.13	&98.47\\
         & BiFuse++ \cite{wang2022bifuse++} &TPAMI'22 &52.49 &0.1123	&0.0915	&0.4853		&88.12	&96.56	&98.69\\
         & UniFuse \cite{jiang2021unifuse} &RAL'21 &50.48 &0.1144	&0.0936	&0.4835	&87.85	&96.59	&98.73\\
         & OmniFusion \cite{li2022omnifusion} &CVPR'22 &42.46 &0.1161	&0.1007	&0.4931		&87.72	&96.15	&98.44\\
         & HRDFuse \cite{ai2023hrdfuse} &CVPR'23 &46.31 &0.1172	&0.0971	&0.5025		&86.74	&96.17	&98.49\\
         & Elite360D \cite{ai2024elite360d} &CVPR'24 &25.54 &0.1115	&0.0914	&0.4875		&88.15	&96.46	&98.74\\
        & SGFormer \cite{zhang2025sgformer} &TCSVT'25 &- &0.1039 &0.0865	&0.4790	&89.46	&96.42	&98.59\\   
        & {Ours} &- &67.84 &{\textbf{0.0955}}	&{\textbf{0.0706}}	&{\textbf{0.4163}}		&{\textbf{90.78}}	&{\textbf{97.11}}	&\textbf{98.88}\\
        
        \midrule
        
        \multirow{6}{*}{\centering S2D3D}
         & EGFormer \cite{yun2023egformer} &ICCV'23 &15.39  &0.1528	&0.1408	&0.4974	&81.85	&93.38	&97.36\\
         & PanoFormer \cite{shen2022panoformer} &ECCV'22 &20.38 &0.1122 &0.0786 &0.3945 &88.74	&95.84	&98.59\\
         & OmniFusion \cite{li2022omnifusion} &CVPR'22 &42.46 &0.1154	&0.0775	&0.3809	&86.74	&96.03	&98.71\\
         & UniFuse \cite{jiang2021unifuse} &RAL'21 &50.48 &0.1124	&0.0709	&0.3555	&87.06	&97.04	&98.99\\
         & Elite360D \cite{ai2024elite360d} &CVPR'24 &25.54 &0.1182	&0.0728	&0.3756	&88.72	&96.84	&98.92\\
        & SGFormer \cite{zhang2025sgformer} &TCSVT'25 &- &\textbf{0.1040} &\textbf{0.0581}	&\textbf{0.3406}	&\textbf{89.98}	&96.93	&99.08\\        
        & {Ours} &- &67.84 &{0.1211}	&{0.0740}	&{0.4042}	&{85.66}	&{\textbf{97.09}}	&{\textbf{99.13}}\\
        \bottomrule
    \end{tabular}
    }
    \label{tab:quantitative_comparison_clip}
\end{table*}

\begin{figure*}[t!]
\centering
 \includegraphics[width=1\linewidth]{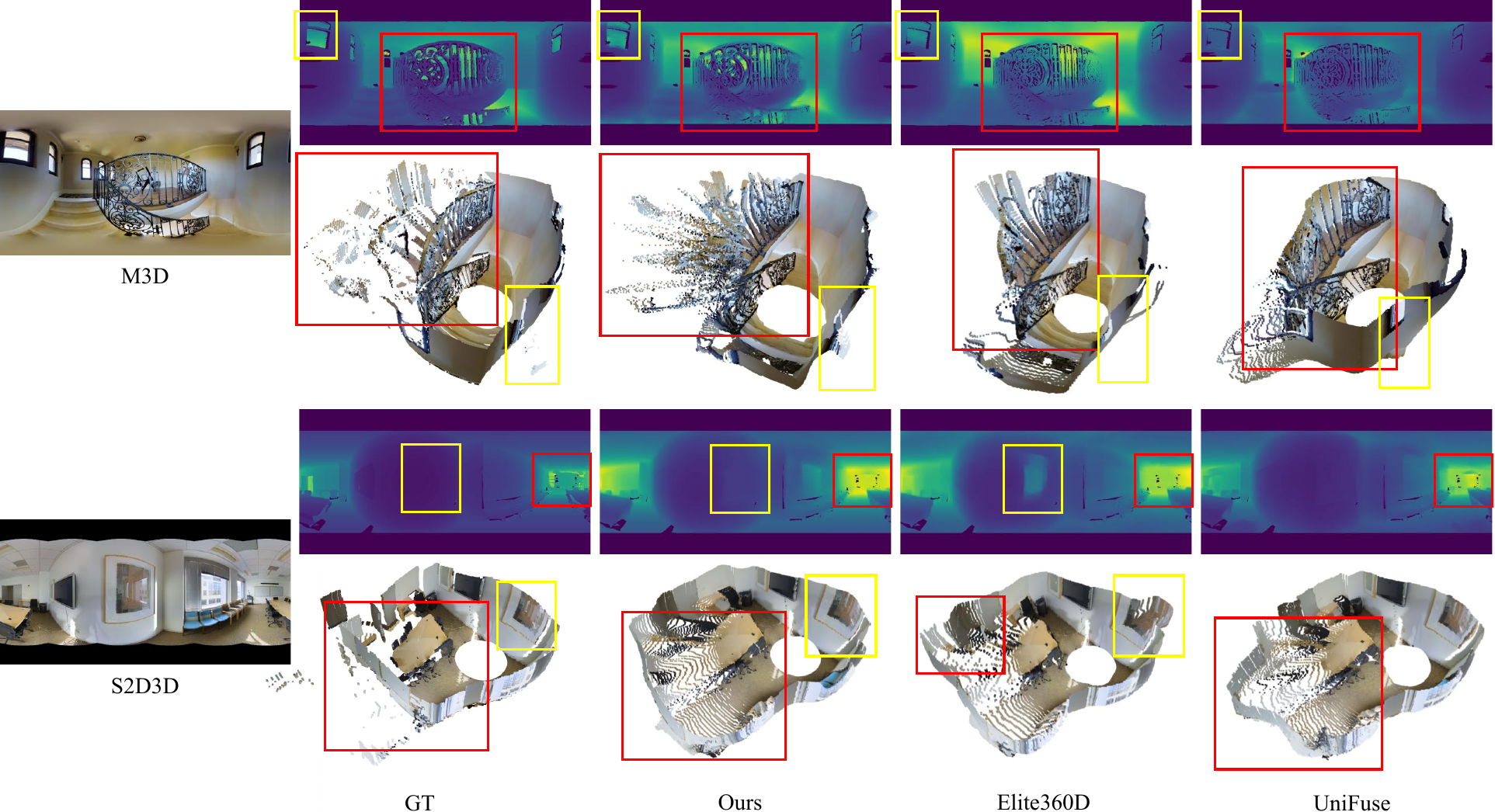}
 \caption{Qualitative comparison results on Matterport3D and Stanford2D3D datasets. We compare our predicted depth maps against Elite360D \cite{ai2024elite360d} and UniFuse \cite{jiang2021unifuse}, highlighting key differences with colored rectangles. The results are visualized as 2D images and 3D point clouds, incorporating both RGB data and depth maps for a more comprehensive analysis.}
 \label{fig:rw_results}
\end{figure*}

\begin{table*}[t]
    \centering
    \caption{Quantitative comparisons against SOTA methods on synthetic datasets. Inputs cover the entire FoV of the scene.}
    \resizebox{\textwidth}{!}{
    \begin{tabular}{c|l|c|c|ccc|ccc}
        \toprule
        \multirow{1}{*}{\centering Dataset} & \multirow{1}{*}{Method}& \multirow{1}{*}{Pub'Year}& \multirow{1}{*}{Params (M)}
        &Abs Rel $\downarrow$  & Sq Rel $\downarrow$ & RMSE $\downarrow$
        & $\delta_1$(\%) $\uparrow$& $\delta_2$(\%) $\uparrow$& $\delta_3$(\%) $\uparrow$\\
        \midrule

        \multirow{7}{*}{\centering Struct3D}
         & UniFuse \cite{jiang2021unifuse} &RAL'21 &30.26 &0.0535	&0.0279	&0.1365	&96.89	&98.76	&99.27\\
         & PanoFormer \cite{shen2022panoformer} &ECCV'22 &20.38 &0.0551 &0.0226 &0.1361 &96.70	&98.80	&99.34\\
         & HRDFuse \cite{ai2023hrdfuse} &CVPR'23 &46.31 &0.0575	&0.0299	&0.1493	&96.36	&98.54	&99.16\\
         & GLPanoDepth \cite{bai2024glpanodepth} &TIP'24 &120.32 &0.0520	&0.0219	&0.1267	&96.86	&98.82	&99.35\\
         & DepthAnything \cite{depthanything} &CVPR'24 &343.04 &0.0825	&0.0382	&0.1969	&94.07	&97.91	&98.89\\
         & ASNGeo \cite{long2024adaptive} &TPAMI'24 &89.56 &0.0802	&0.0332	&0.1862	&94.41	&98.26	&99.13\\         
        & {Ours} &- &67.84 &{\textbf{0.0361}}	&{\textbf{0.0127}}	&{\textbf{0.1030}}	&{\textbf{97.88}}	&{\textbf{99.25}}	&{\textbf{99.63}}\\
        \midrule
         
        \multirow{7}{*}{\centering 3D60}
         & UniFuse \cite{jiang2021unifuse} &RAL'21 &30.26 &0.0720	&0.0373	&0.3012	&94.51	&98.87	&99.64\\
         & PanoFormer \cite{shen2022panoformer} &ECCV'22 &20.38 &0.0617	&0.0213	&0.2234	&96.65	&99.41	&99.80\\
         & HRDFuse \cite{ai2023hrdfuse} &CVPR'23 &46.31 &0.0729	&0.0333	&0.2911	&94.72	&98.92	&99.62\\
         & GLPanoDepth \cite{bai2024glpanodepth} &TIP'24 &120.32 &0.0673	&0.0266	&0.2535		&96.00	&99.30	&99.79\\
         & DepthAnything \cite{depthanything} &CVPR'24 &343.04 &0.1064	&0.0781	&0.4343	&89.12	&97.26	&98.96\\
         & ASNGeo \cite{long2024adaptive} &TPAMI'24 &89.56 &0.0837	&0.0390	&0.3305		&93.24	&98.68	&99.59\\
        & {Ours} &- &67.84 &{\textbf{0.0526}}	&{\textbf{0.0177}}	&{\textbf{0.2121}}		&{\textbf{97.40}}	&{\textbf{99.55}}	&{\textbf{99.86}}\\
        

        \bottomrule
    \end{tabular}
    }
    \label{tab:quantitative_comparison}
\end{table*}

\begin{figure*}[t!]
\centering
 \includegraphics[width=1\linewidth]{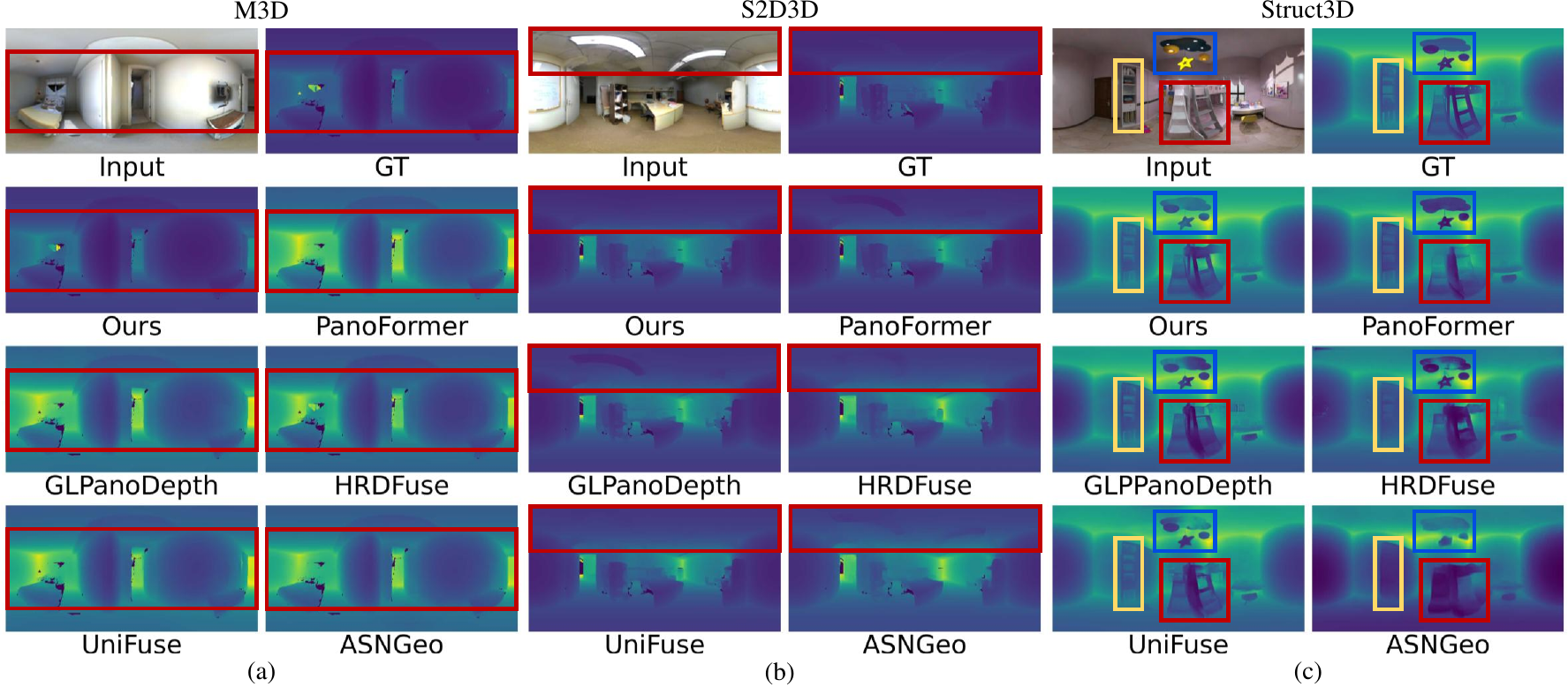}
 \caption{Qualitative comparison results from synthetic datasets (Structured3D and 3D60). We \hk{compare} our predicted results against various methods. All depth maps use a consistent visualization style, and the regions highlighted by rectangles illustrate our model's superior geometric understanding of the scenes. More comparisons can be found in the supplementary materials.}
 \label{fig:3d60_results}
\end{figure*}

\subsection{Comparison with the state-of-the-art}
\subsubsection{\hk{Comparison on Incomplete Real-world Datasets}} 
Table \ref{tab:quantitative_comparison_clip} presents a quantitative comparison on the real-world Matterport3D and Stanford2D3D datasets, using incomplete ERP images as inputs. The results demonstrate that Cross360 achieves superior performance on Matterport3D across all evaluation metrics. For example, with a ResNet-34 backbone, our Cross360 outperforms the most recent state-of-the-art method, Elite360D, with gains of 14.35\% in Abs Rel, 22.76\% in Sq Rel, and 14.61\% in RMSE. On the small-scale Stanford2D3D dataset, Cross360 achieves comparable performance to PanoFormer \cite{shen2022panoformer} and UniFuse \cite{jiang2021unifuse} but leads with the highest accuracy scores, reaching 97.09\% on $\delta_2$ and 99.13\% on $\delta_3$, which reflects our model’s robust accuracy. 
\hk{These results further indicate that, despite being designed to exploit full-sphere context, our method remains highly robust when parts of the ERP are missing. By combining adaptive tangent-plane sampling with masked supervision, Cross360 maintains state-of-the-art accuracy even with the incomplete top and bottom regions commonly found in real-world panoramas, while still benefiting from full 360° coverage when it is available.}
The limited size of the Stanford2D3D dataset constrains our model’s ability to fully capture global contextual relationships, which are essential for our transformer-based network. While other attention-based methods \cite{shen2022panoformer} are less affected by incomplete ERP inputs, our approach is designed for complete depth prediction across all ERP regions, making it more sensitive to missing and highly distorted areas.

Fig. \ref{fig:rw_results} presents qualitative comparisons of depth predictions from our method, Elite360D \cite{ai2024elite360d}, and UniFuse \cite{jiang2021unifuse}, alongside ground truth. The results are visualized as 3D point clouds, with red rectangles highlighting key regions. For the results from Matterport3D, our approach accurately captures hollow areas in structures like handrails and windows (bottom right), whereas other methods misinterpret these regions as solid planes. Similarly, for Stanford2D3D, our model demonstrates a better understanding of overall room geometry, correctly predicting walls and surfaces where Elite360D and UniFuse fail to distinguish structural elements.

\subsubsection{\hk{Comparison on Complete Synthetic Datasets}} 
To further validate our method's performance with complete ERP inputs, we present quantitative results in Table \ref{tab:quantitative_comparison}. As shown, Cross360 significantly outperforms SOTA methods on both the Structured3D and 3D60 datasets. For a fair comparison, we retrain all competing methods using their official hyperparameters, identical data splits, and a consistent range scale (e.g., a maximum depth of 10 meters). We also apply an early-stopping strategy, terminating training at the 15th epoch if no further improvement is observed. For DepthAnything \cite{depthanything}, which is pre-trained on a large-scale dataset, we use its provided pre-trained model and fine-tune it on the 360° datasets, following the same training strategy.

On the Structured3D dataset, a synthetic dataset featuring diverse samples, varied lighting, and accurate ground-truth data, our Cross360 demonstrates exceptional performance across all metrics, achieving improvements over the current SOTA method, GLPanoDepth \cite{bai2024glpanodepth}, with 30.58\% in Abs Rel, 42.01\% in Sq Rel, and 18.71\% in RMSE. Additionally, our model has 43.62\% fewer parameters than GLPanoDepth, making it more efficient. On the 3D60 dataset, Cross360 surpasses the current SOTA method, PanoFormer \cite{shen2022panoformer}, with gains of 14.75\% (Abs Rel), 16.90\% (Sq Rel), and 5.06\% (RMSE).


Notably, depth estimation methods investigated for perspective images, such as DepthAnything \cite{depthanything} and ASNGeo \cite{long2024adaptive}, display significantly higher error metrics when applied to 360° images.
\hk{We include these baselines to illustrate the gap between strong perspective models and methods designed for 360° inputs, following a fair setting that fine-tunes each model on panoramic datasets using its recommended settings. Adding spherical convolutions or spherical positional encodings would require major architectural changes incompatible with the pre-trained weights of DepthAnything and would remove the benefit of its large-scale pretraining. For ASNGeo which is a multi-task network and partially distortion-aware through its depth-to-normal supervision, it still cannot match a model built specifically for 360° geometry.}
\hk{For instance, on the Structured3D dataset our method reduces the Sq Rel error by 66.75\% relative to DepthAnything (0.0127 vs. 0.0382)} and also has the highest parameter count, while ASNGeo achieves a lower $\delta_1$ accuracy of 3.41\% compared to our method.
\hk{These comparisons confirm that even strong perspective-based models remain limited on panoramic data, underscoring the need for architectures tailored to the 360° domain.}

Fig. \ref{fig:3d60_results} presents various qualitative comparison results. Benefiting from aggregated globally coherent features and sufficient local appearance details, without distortion issues, our method demonstrates more accurate geometric details and better adaptation to varying depth ranges. In (a), our method more accurately captures the global structure, as highlighted by the rectangles, including precise estimations of the furthest points and the depth range of the scene. In (b), our model avoids misinterpreting ceiling lights as part of the room structure, demonstrating better semantic understanding. In (c), Cross360 more accurately highlights specific regional structural details, such as star lights, bookcases, and ladders. Additional qualitative results are provided in the supplementary material.

The results highlight Cross360's adaptability and effectiveness, particularly on large-scale datasets like Matterport3D and Structured3D, even with incomplete ERP inputs. The model effectively leverages available information to maximize performance in challenging real-world scenarios, accurately capturing the global structure of scenes.

\subsection{Ablation Study}

\begin{table}[t]
    \centering
    \caption{\rv{Ablation study of individual components (3D60 test set).}}
    \begin{adjustbox}{max width=\columnwidth}
		\begin{tabular}  {l|cc|rr}
			\toprule 
			Method & Abs Rel~$\downarrow$  &RMSE~$\downarrow$ &$\delta_1$~$\uparrow$ & $\delta_2$~$\uparrow$\\
			\midrule
			 Baseline & 0.0656 & 0.2445 & 95.49 & 99.17\\
			 Baseline+CPFA & 0.0533 & 0.2172 & 97.07 & 99.43\\
			 Baseline+PFAA & 0.0576 & 0.2230 & 96.84 & 99.42\\
			 Ours (all together) & 0.0526 & 0.2121 & 97.40 & 99.55\\
			\bottomrule 
	\end{tabular}
    \end{adjustbox}
    \label{tab:module_ablation}
\end{table}
\begin{table}[t]
    \centering
    \caption{\rv{Ablation study on the number of TP patches (3D60 test set).}}
    \begin{adjustbox}{max width=\columnwidth}
		\begin{tabular}  {c|c|c|c|cc|c}
			\toprule 
			Num & FoV & FPS~$\uparrow$ &GFLOPs & Abs Rel~$\downarrow$  &RMSE~$\downarrow$&$\delta_1$~$\uparrow$\\
			\midrule
			 10 & 120 &15.42 &172.12 & 0.0599 & 0.2226 & 96.86\\
			 18 & 90 &11.64 &172.74 & 0.0522 & 0.2148 & 97.21\\
			 \gray26 & \gray72 & \gray9.16 & \gray173.36 & \gray0.0526 & \gray0.2121 &\gray97.40\\
			 46 & 60 &5.78 &174.90 & 0.0505 & 0.2115 & 97.29\\
			\bottomrule 
	\end{tabular}
    \end{adjustbox}
    \label{tab:patch_ablation}
\end{table}

\subsubsection{The effectiveness of modules} We conduct an ablation study to evaluate the effectiveness of the proposed modules in our Cross360. In our experiments, we incrementally add one module at a time to the baseline and use a fixed pre-trained ResNet34 \cite{He_2016_CVPR} model as our ERP-based feature encoder. The baseline consists of two branches in the hierarchical architecture: an ERP branch and a TP branch. The concatenation of these two encoded features, following the TP2ERP geometric transformation, is passed to the corresponding level of the decoder to generate a depth map for multi-level supervision. First, we add the CPFA module to the baseline to obtain cross-projection-aligned features. Next, we combine the baseline with the PFAA module to test the effectiveness of progressively aggregated features with attention to the final prediction. Finally, we integrate both the CPFA and PFAA modules to form the complete architecture.

In Table \ref{tab:module_ablation}, the baseline demonstrates the poorest performance due to its inability to effectively align the ERP and TP representations. Adding either the CPFA or PFAA module to the baseline improves performance, with CPFA showing a more significant enhancement. This indicates that both modules are effective: CPFA uses cross-attention to align features from different domains effectively, while PFAA significantly reduces errors by integrating the most relevant information across scales. When all modules are integrated, performance ultimately improves by 19.82\% in Abs Rel and 13.25\% in RMSE compared to the baseline.

\subsubsection{Non-uniform TP Sampling}
\label{ablation}
The number of non-uniformly sampled TP patches directly influences the accuracy and efficiency of our method. We follow OmniFusion \cite{li2022omnifusion} to determine the number of samples and their corresponding FoVs. This sampling strategy places fewer patches near the poles and more near the equator, compensating for ERP's $1/\cos(\text{latitude})$ pixel density. Unlike a strict six-face cubemap, our design uses intentionally denser, partially overlapping patches to reduce border discontinuities and match equatorial pixel density. This controlled overlap enhances depth estimation by providing smoother feature alignment for the CPFA module's cross-attention mechanism. \rv{As shown in Table \ref{tab:patch_ablation}, our approach demonstrates significantly better performance compared to the cubemap-based UniFuse (in Table \ref{tab:quantitative_comparison}). For instance, our method achieves an RMSE of 0.2226 using 10 patches, whereas UniFuse yields 0.3012 using 6 patches.} This configuration achieves an optimal balance between geometric coverage, computational efficiency, and accuracy, where increasing the number of sampled TP patches improves accuracy at the cost of reduced prediction speed due to increased computational demands.
For any given application of our method, there is therefore a trade-off between accuracy and speed. Sampling the smallest number of 10 TP patches with a 120° FoV yields the lowest accuracy but still enhances performance by 14.75\% in Abs Rel and 5.06\% in RMSE compared to PanoFormer (the current SOTA on 3D60 dataset) and provides the highest FPS (15.42). Sampling 46 patches with a 60° FoV significantly boosts performance by 15.69\% in Abs Rel and 4.99\% in RMSE compared to PanoFormer but results in the lowest FPS (5.78). 
Configurations with 18 and 26 patches offer a reasonable tradeoff. We chose the 26-patch configuration (highlighted in gray in the table) as it shows only a 3.99\% and 0.28\% difference in Abs Rel and RMSE, respectively, compared to the 46-patch setting but with a significantly higher FPS (9.16 versus 5.78).
\hk{To further quantify computational complexity, we additionally report GFLOPs in Table \ref{tab:patch_ablation}. The total cost rises only slightly from 172.12 to 174.90 GFLOPs as the number of patches increases from 10 to 46, representing about a 1.6\% overhead. This small increase shows that the CPFA module adds minor computational cost when sampling more TP patches, and the overall model remains computational complexity efficient as the number of patch sampling grows.}

\subsubsection{\hk{Depth Prediction Accuracy along Distance}} 
\hk{We quantitatively evaluated the accuracy of Cross360’s depth estimation on the 3D60 dataset across five distance intervals, with results up to 10 meters detailed in Table \ref{tab:distance_accuracy}. A gradual decline in accuracy with increasing distance is typical for monocular depth estimation. A key factor in this trend is the severe data imbalance, as the number of pixels at far distances is substantially lower than at near distances, providing fewer valid samples for the model to learn from. 
Cross360's depth estimation accuracy is strongest in the 0–2 m range (containing 74.3\% of pixels). In the 2–6 m range, the performance decline is moderate. Beyond 6 m, the accuracy decreases more noticeably as valid pixels become sparse. This pattern reflects the intrinsic difficulty of estimating depth for distant objects, where panoramic cues weaken and each object occupies fewer pixels. Nevertheless, Cross360 maintains strong performance across all ranges, demonstrating that its spherical-aware feature processing provides competitive accuracy even in challenging long-distance regions.}

\begin{table}[t]
\centering
\caption{Depth Estimation Accuracy along Distance}
\begin{adjustbox}{max width=\columnwidth}
\begin{tabular}{c|c|c|c|c}
\toprule
Distance Range (m)& Distribution & Abs Rel~$\downarrow$& RMSE~$\downarrow$&  $\delta_1$~$\uparrow$\\
\midrule
$[~0.0-2.0~)$  & 74.3\% & 0.0559 & 0.1279 & 97.84 \\
$[~2.0-4.0~)$  & 18.4\% & 0.0673 & 0.3760 & 95.30 \\
$[~4.0-6.0~)$  & 5.0\% & 0.0788 & 0.5392 & 96.14 \\
$[~6.0-8.0~)$  & 1.4\% & 0.1047 & 1.0877 & 91.72 \\
$[~8.0-10.0~)$  & 0.9\% & 0.1410 & 1.7228 & 88.28 \\
\bottomrule
\end{tabular}
\end{adjustbox}
\label{tab:distance_accuracy}
\end{table}
\section{Conclusion}\label{sec:conclusion}
We propose a novel 360° monocular depth estimation method. To address the distortion in ERP images, we introduce the CPFA module that leverages complementary information from both TP and ERP representations using a cross-attention mechanism. CPFA lets the local, distortion-free TP features gain global context from ERP features, ensuring the learning of well-aligned local-with-global receptive features across both domains. To further integrate necessary information across scales and resolve the discontinuity issues inherent in ERP images, we propose the PFAA module. The PFAA module aggregates the most relevant information between scales using the attention mechanism. Our ablation study demonstrates the effectiveness of the proposed modules, and our experiments against other state-of-the-art methods highlight Cross360's outstanding performance across most of the general public benchmarks.
\textbf{Future work} could explore the potential of our architecture for other tasks, such as surface normal estimation and room layout prediction. Additionally, collecting a complete, high-quality real-world dataset would further enhance the model's generalizability across diverse environments, contributing valuable resources to the 360° vision community. \hk{Extending the framework to outdoor panoramic scenes by integrating a semantic segmentation or sky masking module to handle infinite depth regions such as sky and distant backgrounds is another promising direction.}


\bibliographystyle{IEEEtran}
\bibliography{main}

\newpage

\section{Biography Section}
\begin{IEEEbiography}[{\includegraphics[width=1in,height=1.25in,clip,keepaspectratio]{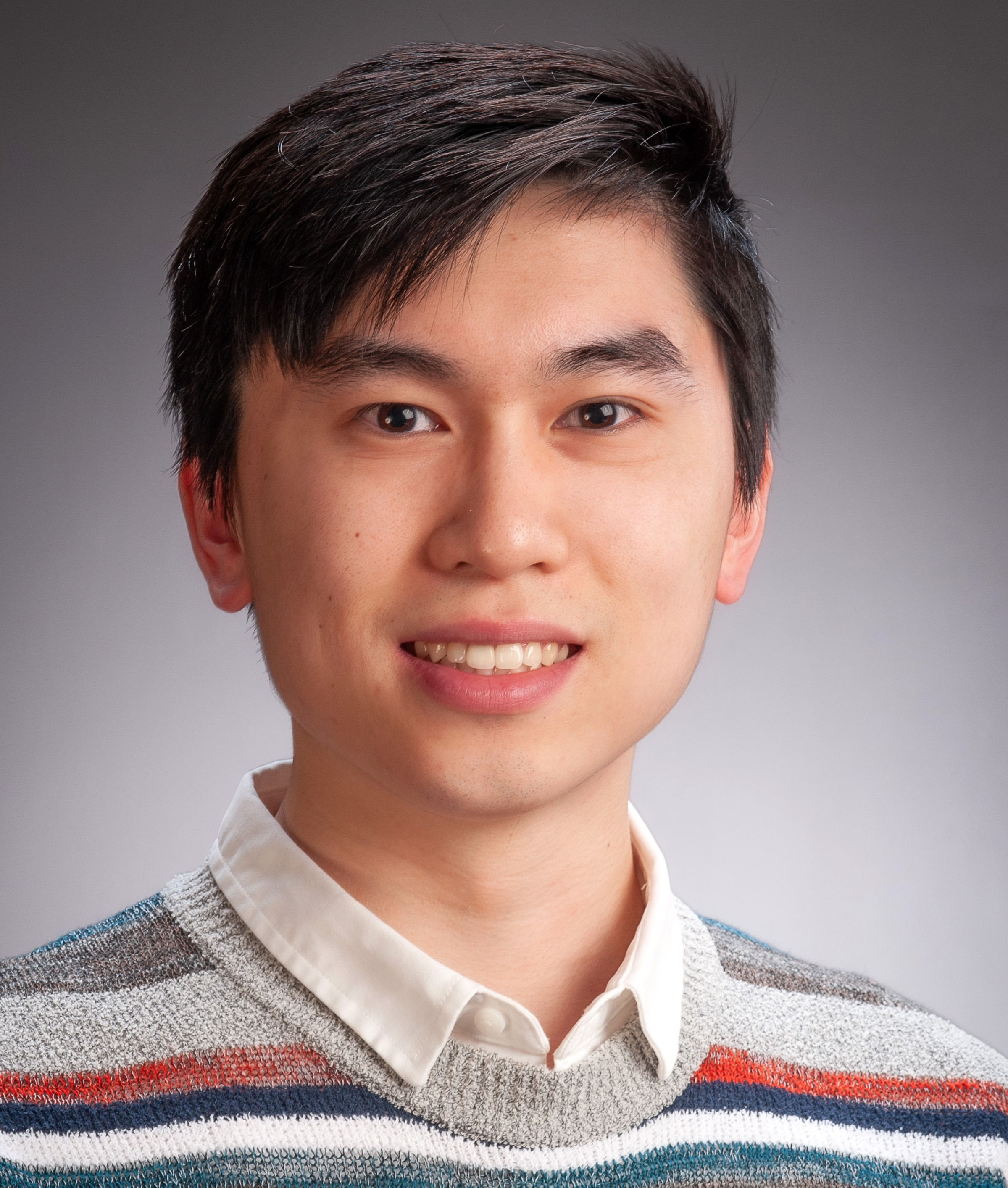}}]{Kun Huang}
received a Ph.D. from Victoria University of Wellington, New Zealand in 2025. His research interests include 360° image and video editing, 3D reconstruction, computer vision, virtual reality and mixed reality. He received a best presentation award from IEEE NZ Central Section Postgraduate Symposium in 2021. He was the student branch chair between 2021-2024 and professional activity coordinator in 2024 of the IEEE New Zealand Central Section.
\end{IEEEbiography}

\vspace{11pt}

\begin{IEEEbiography}[{\includegraphics[width=1in,height=1.25in,clip,keepaspectratio]{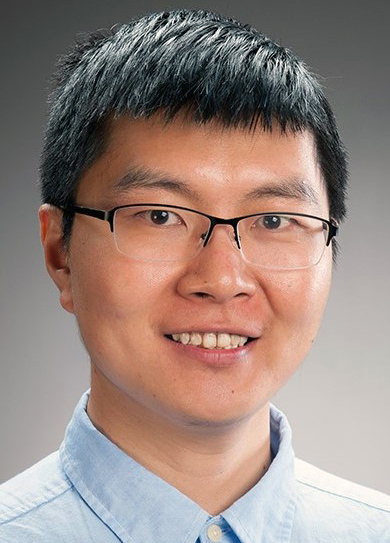}}]{Fang-Lue Zhang}
received a Ph.D. degree from Tsinghua University in 2015. He is currently a Senior Lecturer in Computer Graphics at the Victoria University of Wellington. His research interests include image and video editing, computer vision, and computer graphics. He received a Victoria Early-Career Research Excellence Award in 2019, and a Fast-Start Marsden Grant from the New Zealand Royal Society in 2020. He is on the editorial board of Computers \& Graphics. He served as program chair of Pacific Graphics 2020 \& 2021, and CVM 2024. He is a committee member of IEEE Central New Zealand Section. 
\end{IEEEbiography}

\vspace{11pt}

\begin{IEEEbiography}[{\includegraphics[width=1in,height=1.25in,clip,keepaspectratio]{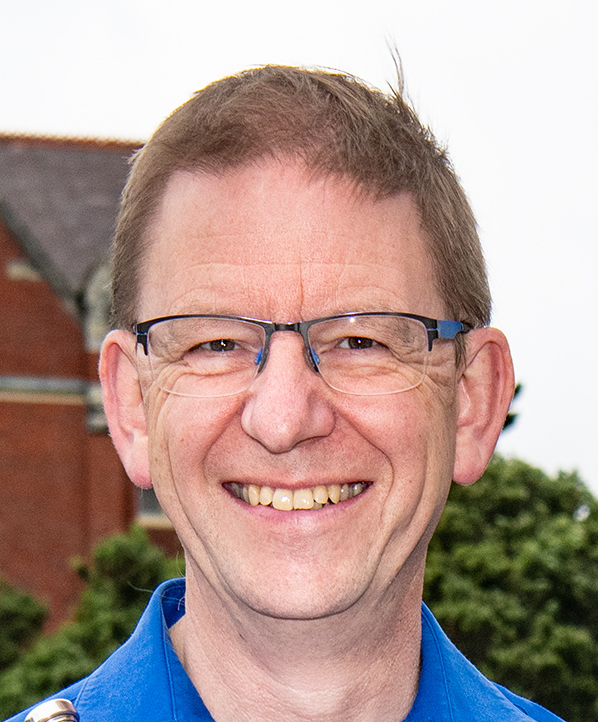}}]{Neil Dodgson}
is Professor of Computer Graphics and Dean of Graduate Research at Victoria University of Wellington. His Ph.D. was in image processing, from the University of Cambridge, where he spent 25 years, becoming full Professor in 2010.  His research is in 3D TV, subdivision surfaces, imaging, and aesthetics. He is a Fellow of Engineering New Zealand, of the Institution of Engineering and Technology  and of the Institute for Mathematics and its Applications.
\end{IEEEbiography}

\vspace{11pt}

\vfill

\end{document}